\documentclass{article}

\usepackage{authblk}
\usepackage{PRIMEarxiv}

\usepackage[utf8]{inputenc} 
\usepackage[T1]{fontenc}    
\usepackage{hyperref}       
\usepackage{url}            
\usepackage{booktabs}       
\usepackage{amsfonts}       
\usepackage{nicefrac}       
\usepackage{microtype}      
\usepackage{lipsum}
\usepackage{amsmath, amsfonts} 
\usepackage{fancyhdr}       
\usepackage{graphicx}       
\graphicspath{{media/}}     
\usepackage{tabularx} 
\usepackage{amsmath}

\usepackage{tcolorbox} 
\usepackage{xcolor} 

\usepackage[backend=biber, style=numeric, citestyle=numeric, maxbibnames=6, minbibnames=6]{biblatex} 
\title{Digital Diagnostics: The Potential of Large Language Models in Recognizing Symptoms of Common Illnesses

}

\author{
  Gaurav Kumar Gupta\textsuperscript{1}, Aditi Singh\textsuperscript{2}, Sijo Valayakkad Manikandan\textsuperscript{3},  Abul Ehtesham\textsuperscript{4}\\
 \textsuperscript{1} Youngstown State University,Youngstown, OH\\
  \textsuperscript{2} Cleveland State University, Cleveland, OH\\
 \textsuperscript{3} NIKE, Inc, Beaverton, Oregon\\
  \textsuperscript{4} The Davey Tree Expert Company, Kent, OH\\
  \texttt{gkgupta@student.ysu.edu,  a.singh22@csuohio.edu, sijopkd@gmail.com, abul.ehtesham@davey.com}
}

\begin{document}
\maketitle

\begin{abstract}
The recent swift development of LLMs like GPT-4, Gemini, and GPT-3.5 offers a transformative opportunity in medicine and healthcare, especially in digital diagnostics. This study evaluates each models diagnostic abilities by interpreting symptoms and determining diagnoses that fit well with common illnesses, and it demonstrates how each of these models could significantly increase diagnostic accuracy and efficiency. Through a series of diagnostic prompts based on symptoms from medical databases, GPT-4 demonstrates higher diagnostic accuracy from its deep and complete history of training on medical data. Meanwhile, Gemini performs with high precision as a critical tool in disease triage, demonstrating its potential to be a reliable model when physicians are trying to make high-risk diagnoses. GPT-3.5, though slightly less advanced, is a good tool for medical diagnostics. This study highlights the need to study LLMs for healthcare and clinical practices with more care and attention, ensuring that any system utilizing LLMs promotes patient privacy and complies with health information privacy laws such as HIPAA compliance, as well as the social consequences that affect the varied individuals in complex healthcare contexts. This study marks the start of a larger future effort to study the various ways in which assigning ethical concerns to LLMs task of learning from human biases could unearth new ways to apply AI in complex medical settings.

\end{abstract}


\section{Introduction}

The development of Large Language Models (LLMs), such as GPT-4, Gemini, and GPT-3.5, has revolutionized natural language processing in the recent past\cite{openai2023gpt4}\cite{geminiteam2024gemini}. With potential applications across multiple domains, these models hold particular promise for medicine in healthcare settings. They possess the ability to parse and produce human-sounding text, which could serve as diagnostic tools. Considering the high volume of patients seeking medical help, digital processes that automate and enhance diagnostic output can boost accessibility and efficiency, ultimately revolutionizing patient care\cite{cui2024multimodal}. However, the adoption of LLMs in diagnosing common, everyday ailments raises the pertinent question about these models’ diagnostic accuracy\cite{ullah2024challenges}. This serves as the premise for this empirical study on the performance of different LLMs in recognizing symptoms of common health conditions. This research investigates the extent to which these models can accurately emulate and diagnose medical illnesses and symptoms, setting the stage for exploring their potential in healthcare applications. LLMs hold promise for modern diagnostic facilities, mirroring these questions regarding their diagnostic accuracy, which stands as a stepping stone to effective healthcare\cite{10.1145/3544548.3581503}. The primary purpose of this study is to appraise and evaluate the performance of the models in diagnosing ailments and to gauge how accurate they are in detecting symptoms of common health conditions. The ultimate aim is to establish whether LLMs can be utilized for digital diagnoses and give us insights into the future of healthcare. The impetus behind this line of inquiry is to scrutinize what processes defense AI entails in this aspect of healthcare. Understanding where these models work (and where they don’t) can essentially guide the refinement of AI technologies into medicine, aiding the broader endeavor of leveraging other forms of technology for healthcare purposes. This study will briefly trace the development of LLMs in healthcare, offer a critical review of relevant literature, and analyze the study’s methodology, setting the stage for a nuanced exploration of the transformative desires of LLMs’ involvement in digital diagnostics, healthcare, and the future of medicine.

\section{Related Work }
\label{sec:headings}

The integration of Large Language Models (LLMs) into healthcare marks a significant transformation in medical diagnostics and patient care \cite{abbasian2024conversational}. These models are fundamentally changing how healthcare providers interact with patients, make decisions, and manage diseases. LLMs enhance the precision and speed of diagnoses, especially in specialties that rely on detailed data analysis such as radiology or pathology, which benefits early detection and treatment planning \cite{10.1145/3582515.3609536}. These models also refine patient interactions by offering personalized health consultations and symptom assessments, increasing both trust and satisfaction. Moreover, LLMs integrate diverse data sources to provide actionable insights, aiding healthcare professionals in making informed decisions \cite{kusa2021prompting}. Their application spans several domains: they support mental health professionals by analyzing speech patterns and behavior, aid chronic disease management through data from wearable technology, and assist in vision care by evaluating imaging for early signs of conditions like diabetic retinopathy \cite{lai2023psyllm}. However, deploying LLMs involves addressing significant challenges such as ensuring data privacy, maintaining model transparency, and managing the ethical implications of automated decision-making in healthcare \cite{humphrey2021privacy}. Despite these hurdles, the potential of LLMs to improve healthcare outcomes is immense, provided there is careful consideration of the technical, ethical, and regulatory issues involved \cite{yu2020artificial} \cite{healthcare11060887}.

\subsection{Advancements in LLMs Capabilities}
The incorporation of Large Language Models (LLMs) into healthcare is ushering in a new era of medical diagnostics and patient care, characterized by significantly enhanced decision-making capabilities and patient interaction. Researchers like Meskó et al.\cite{mesko2017digital} have highlighted the profound impact of LLMs, such as GPT-4, which are adept at processing extensive datasets and delivering responses that are remarkably human-like. These models facilitate a multidimensional approach to healthcare by seamlessly integrating diverse data streams—textual, visual, and sensor-based. This integration enhances diagnostic accuracy and improves patient outcomes by enabling more precise and comprehensive health assessments.

Expanding on the practical applications of these capabilities, Yuan et al.\cite{osti_10448809} have demonstrated how LLMs can optimize the efficiency of clinical trials through enhanced patient-trial matching. Their method utilizes the natural language processing strengths of LLMs to improve compatibility between Electronic Health Records (EHRs) and clinical trial requirements, tackling prevalent issues such as data standardization and patient data privacy. The implementation of a privacy-aware data augmentation strategy has been particularly effective, resulting in a 7.32\% improvement in the matching accuracy of patients to clinical trials. This not only speeds up the process but also ensures greater precision in selecting suitable candidates for trials, thereby accelerating medical research and potential therapeutic discoveries.

In the realm of continuous health monitoring and predictive healthcare, the work of Jin et al.\cite{jin2024healthllm} and Kim et al.\cite{kim2024healthllm} with the Health-LLM system exemplifies the advanced capabilities of LLMs. This system utilizes complex algorithms to analyze data from wearable sensors in real-time, offering personalized health predictions that adapt to the changing parameters of a patient's condition. By integrating and interpreting multifaceted data sets, the Health-LLM system provides targeted advice for chronic disease management, enhancing preventive care and patient adherence to prescribed health regimens.

Moreover, the expansion of LLM applications into mental health care by Xu et al.\cite{10.1145/3643540} represents a significant breakthrough. Their research on Mental-LLM explores the utility of LLMs in analyzing behavioral patterns and textual data to predict mental health outcomes. By employing advanced fine-tuning techniques, these models have shown remarkable proficiency in understanding the nuances of language related to mental health symptoms. This allows for more accurate assessments and personalized treatment plans, addressing the complex needs of mental health patients with unprecedented precision\cite{lai2023psyllm}\cite{10427865}.

These detailed explorations and applications of LLMs across various healthcare domains underscore the immense potential of this technology to revolutionize both the methodology and efficacy of medical care. However, the deployment of such sophisticated AI tools must be managed with a keen awareness of the ethical implications, including patient privacy, data security, and the potential biases inherent in AI models\cite{humphrey2021privacy}. As LLMs continue to evolve and their integration into clinical practice deepens, it is crucial to maintain rigorous standards for data handling and model transparency to fully realize their benefits in a responsible and ethical manner\cite{10.1145/3544548.3581503}.

\subsection{Challenges in Healthcare Integration}
The integration of generative AI and Large Language Models (LLMs) into healthcare systems unveils a spectrum of challenges that are as diverse as they are complex. Yu et al. \cite{yu2020artificial} meticulously detail these challenges, placing a strong emphasis on the critical need for robust data privacy measures, precise model fine-tuning, and thorough system implementation strategies to ensure that AI deployment achieves its intended outcomes without compromising security or efficiency. Their research highlights the indispensable role of collaborative co-design, which involves both clinicians and patients in the development process. This collaborative approach not only ensures that AI tools are finely attuned to specific medical needs but also mitigates risks related to data security and breaches of patient privacy, which are paramount concerns in digital healthcare\cite{humphrey2021privacy}.

Expanding on these foundational issues, Singh et al. \cite{Singh2024} address the psychological implications of AI integration within clinical settings, particularly focusing on the cognitive biases that LLMs might exhibit, such as overconfidence and underestimation. These biases have significant ramifications for clinical decision-making, potentially leading to diagnostic errors or misguided patient care strategies. The authors advocate for the establishment of advanced mechanisms that can assess and adjust the confidence levels of AI systems. Ensuring that LLM outputs are rigorously scrutinized and verified by medical professionals is crucial to maintaining the integrity and trustworthiness of AI-assisted diagnostics.

In a similar vein, Ullah et al. \cite{ullah2024challenges} delve deeper into the technical and interpretative challenges that arise when LLMs are applied to diagnostic medicine, especially in fields requiring nuanced analysis like digital pathology. They identify key obstacles such as the absence of deep contextual understanding and the interpretability issues of complex medical data processed by LLMs. Moreover, biases embedded in the training datasets can lead to skewed AI diagnostics, reflecting the "black-box" nature of these systems. This opacity in AI decision-making processes poses significant hurdles to clinical acceptance and reliability\cite{10.1145/3544548.3581503}.

Furthermore, the Health-LLM system, discussed by Kim et al. \cite{kim2024healthllm}, serves as a case study for the practical challenges associated with deploying LLMs in a healthcare setting focused on leveraging wearable sensor data for health predictions. While the system demonstrates considerable potential for enhancing personalized healthcare through sophisticated data analysis, it also highlights significant technical challenges in accurately processing and interpreting multimodal, domain-specific data. These challenges necessitate continuous advancements in AI technologies to improve their accuracy, reliability, and applicability in diverse clinical environments\cite{ghosh2023clipsyntel}.

Lastly, Montagna et al. \cite{10.1145/3582515.3609536} highlight the logistical and ethical challenges in implementing LLM-based chatbot systems for managing chronic diseases. Their focus is on the intricacies of maintaining data privacy within decentralized architectures, crucial for the secure handling of sensitive patient data. They propose an architectural framework that not only supports chronic disease management but is also adaptable to various medical conditions and compliant with strict privacy regulations. However, the practical deployment of such systems faces significant challenges, including the reliability of AI applications in medical settings, the imperative for extensive clinical trials, and ongoing concerns about the confidentiality and security of patient data\cite{baricparker2020patient}.

Addressing these multidimensional challenges is essential for the successful integration of LLMs into healthcare. It requires a concerted effort to enhance technical capabilities, develop ethical AI practices, and enforce strict regulatory standards\cite{yuan2021improving}. By tackling these issues head-on, the medical community can better harness the transformative potential of AI, ultimately leading to improved patient outcomes and more efficient healthcare delivery systems.

\subsection{LLMs Integration in Healthcare Diagnostics}
The research highlighted in this review underscores the dual nature of integrating Large Language Models (LLMs) into healthcare diagnostics. Meskó et al. \cite{mesko2017digital} and Kusa et al. \cite{kusa2021prompting} provide foundational perspectives on both the potential and pitfalls of employing LLMs in medical contexts. Meskó et al. illuminate the extensive capabilities of LLMs, particularly their ability to manage and synthesize vast amounts of multimodal data—combining text, images, and possibly real-time data—to enhance diagnostic accuracy and patient outcomes. Their discussion extends beyond traditional data handling and explores how these models can adapt to the nuanced demands of healthcare by offering personalized diagnostic insights and treatment options.

On the other hand, Kusa et al. delve into the challenges that arise due to the sensitivity of LLMs to user input variations. Their detailed analysis reveals how differing symptom descriptions and entrenched patient beliefs can skew the outcomes of health diagnostics. The study highlights the critical need for systems that can discern and adjust to such variations, thereby avoiding potential misdiagnoses or errors that could arise from automated processes. This sensitivity to input underscores a significant challenge: the LLMs’ reliance on the quality and nature of the data fed into them, which can significantly impact their effectiveness and reliability.

The insights from these studies are instrumental for ongoing research aimed at evaluating the diagnostic accuracy of LLMs in healthcare. They not only underscore the immense potential of these technologies to transform medical diagnostics through enhanced data processing capabilities but also highlight the complexities involved in their implementation. The integration of LLMs into clinical settings demands the development of robust methodologies that can effectively handle the variability in user inputs and the inherent sensitivities of the models\cite{yu2020artificial}.

The effective deployment of LLMs in healthcare diagnostics depends crucially on their ability to provide accurate, reliable medical advice that aligns with professional standards. This necessitates a balanced approach to technology integration, where the benefits of LLM capabilities in processing and analyzing complex datasets are harmonized with the need for precision and reliability in medical diagnostics\cite{Singh2024}.

 While the promise of LLMs in revolutionizing healthcare diagnostics is undeniable, the path to their full integration is layered with technical and ethical challenges. It requires a strategic and meticulous approach to ensure that the deployment of these technologies enhances, rather than complicates, the diagnostic process. The insights provided by Meskó et al. \cite{mesko2017digital} and Kusa et al. \cite{kusa2021prompting} are crucial in guiding this integration, advocating for a careful and informed application of LLMs in healthcare to truly harness their potential in improving patient care and treatment outcomes. This calls for continued research, development, and refinement of LLM technologies to meet the stringent demands of healthcare applications, ensuring that the future of medical diagnostics is both innovative and ethically sound.

\section{Methodology }

\subsection{Description of the LLMs Evaluated}

This study assesses the diagnostic precision of three user-friendly Large Language Models (LLMs) likely to be most useful in clinical settings – GPT-4, Gemini, and GPT-3.5. Each of these models has been shown to be adept at generating novel and clinically useful information, and able to perform not only clinical prediction and diagnosis, but also aiding in the maintenance and recovery of health through data-driven insights. These models are easily accessible, widely used by the general public, and demand rapid evaluation for their suitability and usefulness in clinical tasks and healthcare research. Each LLMs model has a unique computational strength in working with clinical datasets, and all three stand to improve diagnostic reliability in healthcare settings\cite{kusa2021prompting}.

\textbf{GPT-4:} Developed by OpenAI, GPT-4 stands as a leading entity in the realm of language understanding and generation. This model is distinguished by its advanced capability to interpret complex inquiries. Notably, its architectural design is optimally aligned for the assessment of diagnostic accuracy based on descriptions of medical symptoms. Demonstrating profound efficacy within the medical sphere, GPT-4 has attained a remarkable accuracy rate of 75 percent on the Medical Knowledge Self-Assessment Program. This achievement not only highlights GPT-4's intricate understanding of complex medical queries but also accentuates its critical role in enhancing diagnostic precision from symptom narratives \cite{openai2023gpt4}.

\textbf{Gemini:} Gemini represents a significant advancement in the landscape of Large Language Models (LLMs), particularly with its tailored design for domain-specific applications, including healthcare. The architecture of Gemini is meticulously crafted to enhance the understanding and generation of nuanced responses within these specialized domains. This makes Gemini an invaluable tool for a wide range of tasks, notably in medical diagnostics and healthcare inquiries, where precision and accuracy are paramount. Its ability to seamlessly integrate and reason across multimodal inputs further underscores its potential to revolutionize how medical information is processed and interpreted, setting a new benchmark for AI in healthcare\cite{geminiteam2024gemini}.

\textbf{GPT-3.5:} Serving as a foundational benchmark, GPT-3.5, the precursor to GPT-4, possesses substantial capabilities in language understanding and generation, albeit with slightly lesser proficiency compared to its successor. Its utilization furnishes a comparative framework to evaluate progress in Large Language Models (LLMs) and their applicability in medical diagnostics. Despite its predecessor status, GPT-3.5 has demonstrated commendable performance in the medical domain, notably achieving a 53 percent accuracy rate on the Medical Knowledge Self-Assessment Program. This metric underscores its capability to process and understand clinical and healthcare-related inquiries, thus marking a significant step towards leveraging AI in the enhancement of diagnostic accuracy \cite{openai2023gpt4}.

\begin{figure}[h]
    \centering
    \includegraphics[width=\textwidth]{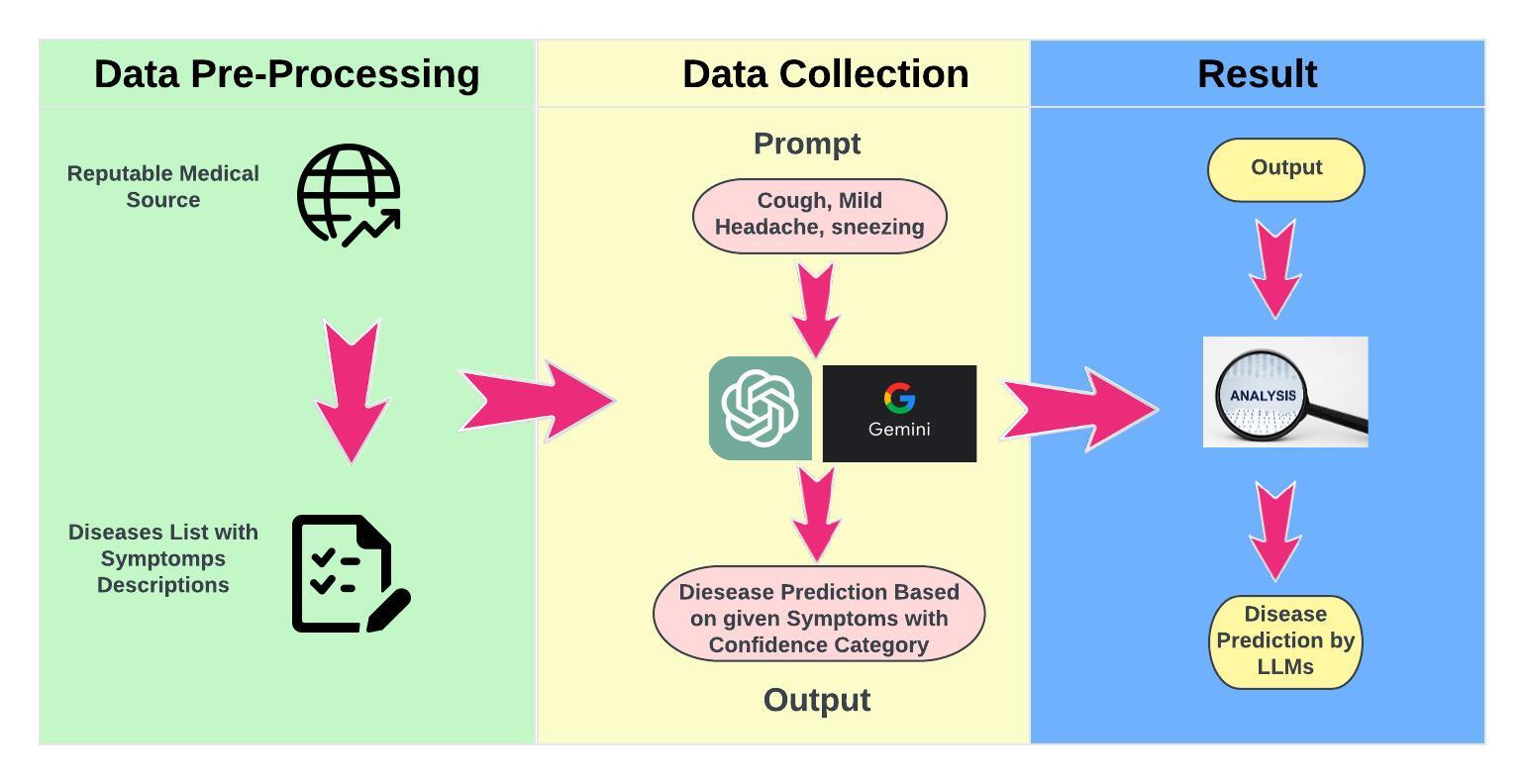}
    \caption{Data Collection Process}
    \label{fig:Data Collection Process}
\end{figure}

\subsection{Data collection methods:}
The foundational dataset for this study was constructed from data extracted from reputable medical sources, including the Centers for Disease Control and Prevention (CDC), World Health Organization (WHO), Mayo Clinic, Cleveland Clinic, and Johns Hopkins Hospital. The selection criterion for diseases focused on conditions frequently encountered in daily life to evaluate the potential utility of Large Language Models (LLMs) in providing diagnostic insights for common ailments. Seasonal allergies, the common cold, and food-related issues such as allergies or diarrhea were included due to their prevalence among the general population, while chronic and complex diseases like cancer were deliberately excluded due to ethical considerations and the current limitations of LLMs in handling such diagnoses accurately.
For each selected disease, a detailed list of symptoms was compiled, forming a comprehensive dataset that pairs these symptoms with their corresponding disease names. This data was organized to support the creation of diagnostic prompts, designed to query disease predictions based on symptom descriptions.
These diagnostic prompts were meticulously crafted to present the symptoms of each disease, asking for disease predictions along with a confidence score for each diagnosis. The prompts were uniformly applied to maintain consistency in the evaluation process. Subsequently, the responses were manually verified to assess the accuracy and reliability of disease prediction, grounding the study's findings in a robust methodological framework.
This methodology underscores the study's aim to investigate the applicability of LLMs as a supportive tool for individuals in recognizing common health conditions. By focusing on diseases frequently encountered, this research contributes valuable insights into the capabilities and limitations of AI technologies in everyday health applications.

\newenvironment{chat}{
  \begin{center} 
   \vspace{-5mm} 
  \begin{tcolorbox}[
    colback=blue!5, 
    colframe=blue!60!black, 
    width=0.95\textwidth, 
    height=5cm, 
    enlarge top by=10mm, 
    enlarge bottom by=10mm, 
    sharp corners,
    boxrule=0.5pt,
    top=2pt,
    bottom=2pt,
    right=2pt,
    left=2pt,
    boxsep=2pt
  ]
  \LARGE 
  \color{red} 
}{
  \end{tcolorbox}
  \end{center}
}

\noindent\textbf{Prompt for models:} \\
The following dialogue presents a prompt used to test the diagnostic capabilities of various language models.
\begin{chat}
 Based on these symptoms: \textcolor{blue}{[Symptoms]}, identify the only one disease based on the symptoms that match closely. Provide me confidence level—\textcolor{blue}{Low}, \textcolor{blue}{Medium}, \textcolor{blue}{High}—to each, based on how closely the symptoms align with the diseases you have predicted based on the {Symptoms}. No explanation needed, provide me exact one-word disease name.
\end{chat}

\subsection{Evaluation Metrics for Diagnosing Diseases through LLMs}

The efficacy of Language Learning Models (LLMs) in diagnosing diseases from descriptions of medical symptoms was evaluated using a detailed, multi-step manual process. This approach incorporated the use of precision, recall, and the F1 score, metrics renowned for their ability to offer a rounded perspective on the accuracy of predictive models in both identifying correct diagnoses and highlighting the omission of relevant diagnoses.

In our study, we scrutinized the LLMs’ outputs for each dataset entry, systematically classifying each response based on its diagnostic accuracy. The classifications were as follows:

\begin{itemize}
    \item \textbf{True Positive (TP):} Instances where the LLM correctly identified the disease, showcasing the model’s capability to accurately match symptom descriptions with the correct disease diagnosis.
    \item \textbf{False Positive (FP):} Instances where the LLM incorrectly identified a disease, attributing a condition to the symptom descriptions that did not align with the actual disease present, thereby overestimating the model's diagnostic accuracy.
    \item \textbf{False Negative (FN):} Instances where the LLM either attributed a different disease than the one actually present based on the symptom descriptions or failed to recognize the presence of a disease altogether, thus underestimating the model's diagnostic sensitivity.
\end{itemize}

We then proceeded to compute the following metrics:

\begin{itemize}
    \item \textbf{Precision:} This metric evaluates the exactness of the model’s positive predictions (i.e., the proportion of TP observations among all positive diagnoses made by the model), offering insight into the accuracy of the model's disease identification.
    \begin{equation}
    \text{Precision} = \frac{TP}{TP + FP}
    \label{eq:precision}
    \end{equation}

    \item \textbf{Recall:} This metric assesses the model’s ability to identify all pertinent instances (i.e., the ratio of TP observations to all actual positives within the dataset), providing a measure of the model’s comprehensiveness in disease detection.
    \begin{equation}
    \text{Recall} = \frac{TP}{TP + FN} \
    \label{eq:recall}
    \end{equation}
    
    \item \textbf{F1 Score:} This metric serves as a balanced measure of both precision and recall, particularly valuable when the contributions of both metrics are of equal importance. It is calculated as the harmonic mean of precision and recall, furnishing a singular measure of the model’s overall diagnostic performance.
    \begin{equation}
    \text{F1 Score} = 2 \cdot \frac{\text{Precision} \cdot \text{Recall}}{\text{Precision} + \text{Recall}}
    \label{eq:f1score}
    \end{equation}

\end{itemize}

Employing these metrics enables a comprehensive evaluation of the LLMs' diagnostic capabilities, providing nuanced insights into the precision of correct diagnoses and the models’ overall efficacy in disease identification.

\subsection{Experimental Setup}

The experimental setup involved an intricate process where symptom-based prompts, meticulously crafted from our curated dataset, were presented to the LLMs under study. Each model was subjected to an identical sequence of prompts, ensuring that the variability in responses could solely be attributed to the models’ interpretative capabilities. This methodical approach facilitated a standardized comparison of query presentation and the verbatim recording of responses. The recorded predictions were then methodically compared against the actual diseases linked with the given symptom descriptions. This comparison allowed for an evaluative analysis of each model’s diagnostic accuracy, providing a clear framework for assessing performance within a controlled and consistent environment. This level of detail in the experimental setup was crucial for ensuring the reliability and validity of our findings, underscoring the study's contribution to understanding the potential of LLMs in medical diagnostics.

\begin{table}[ht!]
 \caption{Model Performance and Characteristics}
  \centering
  \resizebox{\textwidth}{!}{
  \begin{tabular}{p{1.5cm}p{2.5cm}p{2cm}p{3.5cm}p{3cm}p{1.5cm}p{1cm}p{1cm}}
    \toprule
    Model & Number of Parameters & Training Data Size & Primary Application Areas & Unique Features & Precision & Recall & F1 Score \\
    \midrule
    GPT-4 & 175 billion & >800 GB & General-purpose language tasks, content creation, question-answering & Improved context understanding, Multitasking capabilities & 0.95 & 0.91 & 0.92 \\
    GPT-3.5 & 13 billion & ~570 GB & Text completion, language translation, coding assistance & Fine-tuning capabilities for specific tasks & 0.90 & 0.84 & 0.86 \\
    Gemini & 50 million & ~100 GB & Niche applications (e.g., medical, legal), Low-resource languages & Optimized for efficiency and specific domain knowledge & 0.97 & 0.69 & 0.80 \\
    \bottomrule
  \end{tabular}}
  \label{tab:model_performance}
\end{table}

\section{Results}

\subsection{Overview of Findings}

Our comprehensive evaluation delved into the diagnostic abilities of three state-of-the-art Language Learning Models (LLMs) --- GPT-4, GPT-3.5, and Gemini. The goal was to assess how effectively these models can analyze and diagnose medical conditions based on detailed descriptions of symptoms. The findings, illustrated in Figure 1, reveal significant differences in the diagnostic accuracy and capabilities of each model, shedding light on their potential utility in clinical settings. GPT-4 emerged as the standout performer in our study, demonstrating exceptional diagnostic accuracy. This model's success is attributable to its extensive training on a vast array of medical literature and patient data, which has equipped it with a profound understanding of medical symptomatology. GPT-4’s ability to consistently and accurately identify diseases from symptom descriptions showcases its advanced algorithmic structure and sophisticated data processing capabilities. It sets a benchmark in the realm of AI-driven medical diagnostics, proving to be a robust tool that could revolutionize how healthcare providers approach diagnosis and treatment planning. Close behind in terms of performance, GPT-3.5 displayed robust diagnostic skills as well. Although it did not surpass GPT-4, its effectiveness in converting complex symptom data into accurate health assessments makes it a valuable asset in the medical field. GPT-3.5 supports clinical decision-making by providing reliable interpretations of medical contexts, which can greatly aid physicians in diagnosing and understanding patient conditions more effectively. Its solid performance underlines the reliability of well-trained LLMs in handling medical diagnostic tasks, highlighting the potential for AI to assist significantly in everyday healthcare operations. Gemini, although it did not achieve as many correct diagnoses as its counterparts, was noted for its extraordinary diagnostic precision. This model adopts a conservative approach, prioritizing accuracy over quantity in its outputs. Such high-confidence predictions make Gemini especially suitable for use in clinical scenarios where accuracy is critical, such as in the diagnosis of complex or rare conditions where the cost of a misdiagnosis can be particularly high. Gemini’s approach to minimizing false positives is vital in clinical practices where precision is paramount, and the margin for error is minimal. The collective performance of these LLMs paints an optimistic picture of the role of AI in enhancing medical diagnostics. The integration of these advanced models into healthcare could lead to faster, more accurate, and highly reliable diagnostic processes. GPT-4's broad diagnostic capabilities, GPT-3.5’s dependable performance, and Gemini’s meticulous precision collectively embody the advancement of artificial intelligence within the healthcare sector. These findings from our study not only underscore the substantial progress that AI technology has made in unraveling and understanding the complexities of human health but also pave the way for their future applications in medical practice. By enhancing the efficiency and accuracy of diagnostics, these models could serve as invaluable tools for medical practitioners, enabling better patient outcomes and transforming the landscape of healthcare delivery. As we continue to explore and refine these technologies, their integration into clinical workflows holds the promise of making healthcare more effective, personalized, and accessible to all.

\begin{figure}[h]
    \centering
    \includegraphics[width=\textwidth]{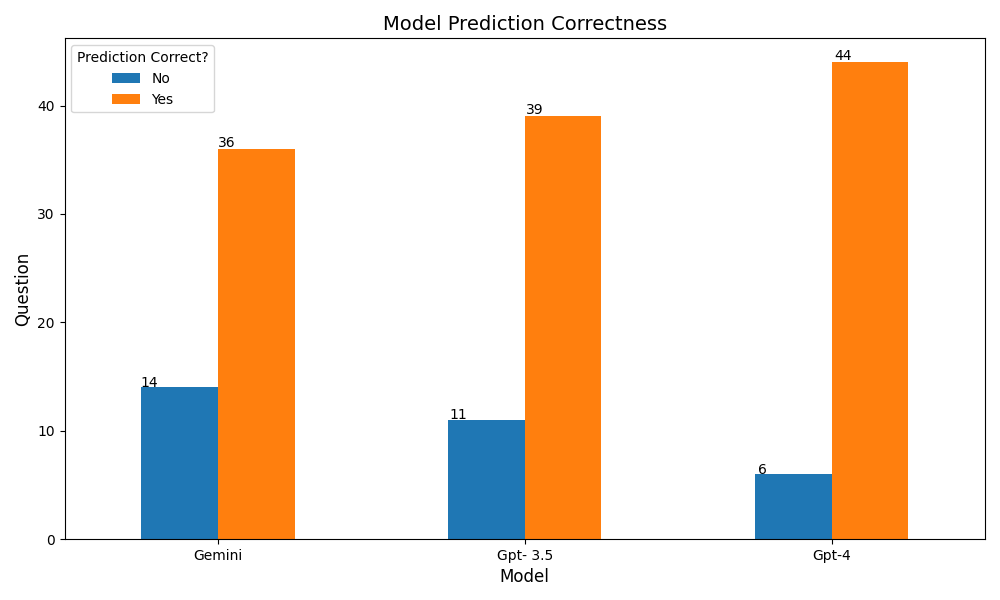}
    \caption{Model Prediction Correctness}
    \label{fig:model_prediction_correctness}
\end{figure} 

\subsection{Comparative Analysis}
The comparative evaluation of GPT-4, GPT-3.5, and Gemini through our study provides an illuminating overview of their diagnostic abilities, each underscored by unique strengths as revealed by the performance metrics summarized in Table 2 and visually elucidated in Figure 2. GPT-4, with its outstanding number of correct answers, stands as a testament to its comprehensive training regimen, encompassing a wide array of medical data. This extensive preparation is reflected in its superior F1 score, indicative of the model's proficiency in deciphering complex medical language and accurately mapping symptoms to diagnoses.

Figure 2 graphically displays the comparative accuracy of the models, highlighting GPT-4’s dominance in correctly answering questions, thereby showcasing its exceptional capability to navigate the complexities inherent in the symptom-diagnosis correlation. This visual representation provides a clear perspective on each model's performance, with GPT-4 leading in precision and understanding.

Further enhancing our analytical perspective, Figure 3 delves into the confidence levels associated with each model's predictions. Here, the confidence distributions of GPT-4 and GPT-3.5 are primarily classified under the 'High' category, underscoring their robust assertiveness and reliability in diagnostic conclusions. In stark contrast, Gemini's inclination towards high-confidence responses, despite a lower overall number of predictions, spotlights its unparalleled precision. This trait is particularly crucial in healthcare contexts where the stakes of misdiagnosis are high, emphasizing the need for accuracy and high confidence in diagnostics.

However, Gemini's impressive precision comes at the cost of recall, as evidenced by its performance metrics. This suggests a cautious approach to diagnosis, where the model opts for certainty over breadth, potentially overlooking certain conditions in the process. Meanwhile, GPT-3.5, although not as advanced as GPT-4, demonstrates significant diagnostic utility, balancing precision and recall effectively. Its solid performance affirms its value in scenarios where cutting-edge models like GPT-4 might not be available or necessary.

Integrating these insights, the study underscores the multifaceted diagnostic capabilities inherent in these LLMs. GPT-4 emerges as a versatile asset across various medical domains, whereas Gemini's precision earmarks it as an invaluable resource for confirming diagnoses with high confidence. On the other hand, GPT-3.5's reliable performance ensures its continued relevance in the evolving landscape of AI in healthcare.

The nuanced findings from our study advocate for a strategic incorporation of LLMs within healthcare settings, leveraging each model's distinct strengths. Such an approach not only augments the diagnostic process but also enhances the overall quality of care, paving the way for an AI-integrated healthcare ecosystem that prioritizes accuracy, efficiency, and patient safety. This comprehensive analysis and the subsequent recommendations aim to foster a deeper understanding of how LLMs can be optimally utilized in healthcare, marking a significant step towards the realization of AI's potential in improving diagnostic outcomes and patient care.

\begin{table}[ht]
\centering
\caption{Comparative Performance of LLMs in Digital Diagnostics}
\begin{tabular}{lccc}
\toprule
Model   & Precision & Recall & F1 Score \\
\midrule
GPT-4   & 0.95      & 0.91   & 0.92     \\
Gemini  & 0.97      & 0.69   & 0.80     \\
GPT-3.5 & 0.90      & 0.84   & 0.86     \\
\bottomrule
\end{tabular}
\label{tab:model_performance}
\end{table}

\begin{figure}[h]
    \centering
    \includegraphics[width=\textwidth]{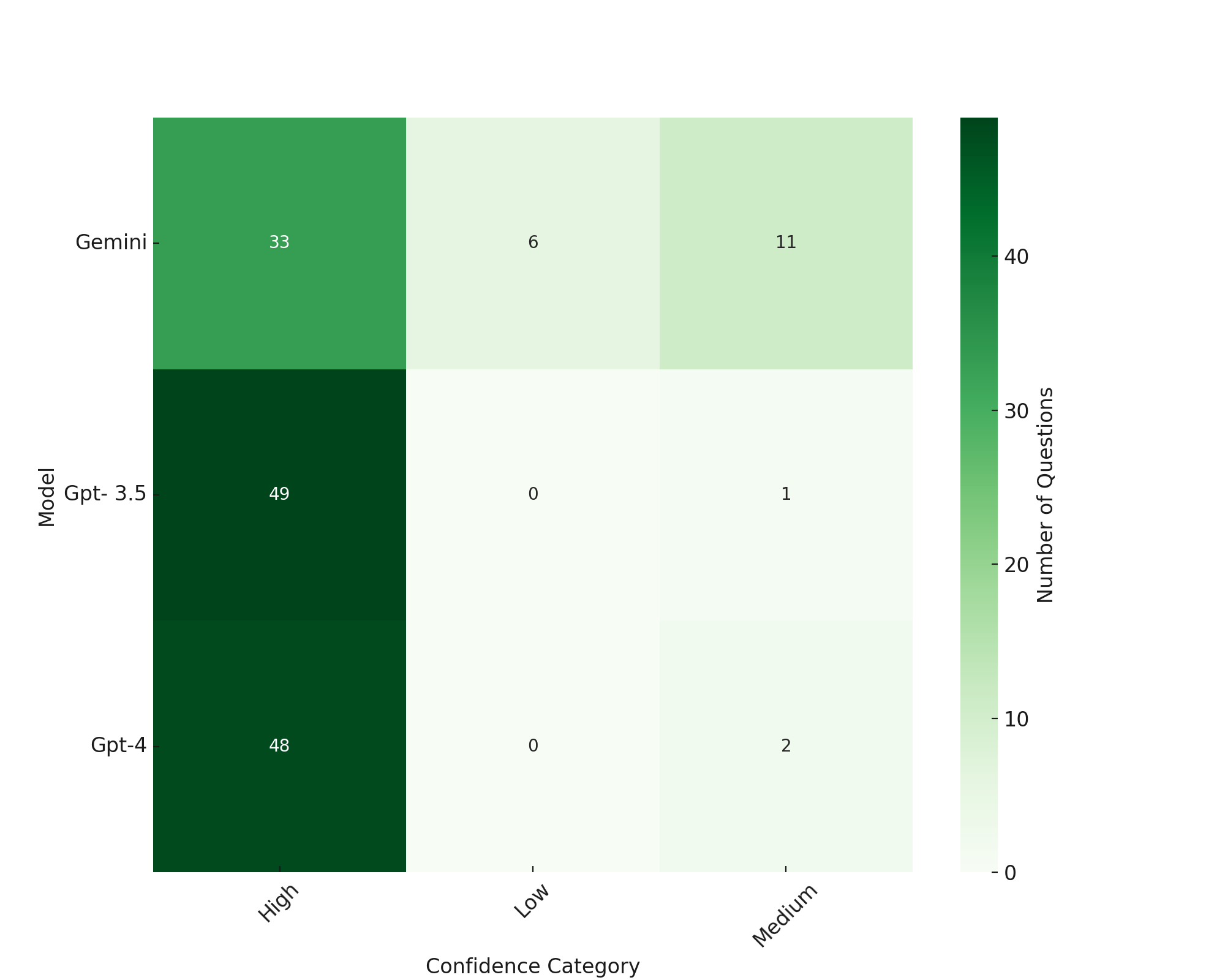}
    \caption{Model Confidence Category}
    \label{fig:model_prediction_correctness}
\end{figure}

\subsection{Concluding Insights}

This study offers insights into the diagnostic capabilities of GPT-4, Gemini, and GPT-3.5. GPT-4 is noted for its high accuracy in diagnosing common illnesses, making it the top performer among the models tested. Gemini, with its high precision, shows great promise as a supplemental tool for digital diagnostics, especially in tasks that require pinpoint accuracy. GPT-3.5, ranking second in disease prediction accuracy, remains a reliable option despite being slightly less advanced. It showcases the rapid advances in LLM technology and its tangible benefits for healthcare.

These findings highlight the substantial potential of integrating LLMs into digital healthcare solutions. They also emphasize the need for ongoing development to improve the accuracy and reliability of these models, ensuring they can effectively meet clinical needs in digital healthcare environments.

\section{Discussion}

\subsection{Interpretation of Results}

This study carefully examined the capabilities of three advanced Language Learning Models (LLMs)—GPT-4, GPT-3.5, and Gemini—in diagnosing everyday illnesses based on symptom descriptions. The standout performer, GPT-4, showcases the potential of AI in medical diagnostics through its ability to understand and process complex medical data. This model's effectiveness highlights its extensive training across diverse medical scenarios, making it exceptionally good at matching symptoms with the correct medical conditions.
Although not as advanced as GPT-4, GPT-3.5 still demonstrated significant capability in making accurate medical diagnoses. Its ability to deliver reliable assessments makes it a useful tool in healthcare, particularly where newer technologies might not yet be available. Gemini, while it ranked third in the number of correct diagnoses, excelled in precision. This high level of accuracy when making a diagnosis shows that Gemini is very careful about ensuring its predictions are correct, which is crucial in medical settings where errors can have serious consequences

\subsection{Enhancing Diagnostic Processes with Large Language Models}
The introduction of Large Language Models (LLMs) into the medical context is likely to change the way healthcare is provided at the very first stages of contact between a patient and a healthcare professional. In the future, LLMs will likely improve the speed and quality of first medical consultations, allowing more rapid assessments of patient information, thereby taking much pressure off medical experts or allowing other professionals to fill the role. In settings where resources are scarce, especially when access to first human consultations is unfeasible, LLMs could quickly analyze the data provided by symptoms and tell the patient what they might be suffering from\cite{healthcare11060887}\cite{10.1145/3544548.3581503}. 

In triaging these cases, LLMs could help to prioritize patients based on urgency and route them to the appropriate level of care. All of this could lead to improved patient outcomes, such as faster interventions, fewer missed appointments, and reduced wait times before treatment. Patient education is also one of the benefits of using these models. LLMs can also be used to provide patients with additional information about their symptoms and potential diagnoses—additional knowledge that empowers people to better understand their health\cite{kusa2021prompting}. 

Yet, such incorporation of AI-enabled tools in healthcare systems is bound to be fraught, and we shall have to navigate several hefty issues. For example, adherence to high healthcare regulations, such as the Health Insurance Portability and Accountability Act (HIPAA) in the United States, this is comprehensive federal legislation that sets metalegal standards and outlines the protection and secure transfer of patients’ data in healthcare systems\cite{humphrey2021privacy}. All the information gathered by LLMs during healthcare interactions must be strictly HIPAA compliant; they should master the vital data and conversation with full security and confidentiality. 

In addition, LLMs might prove valuable, but if integrated, their use must complement, rather than substitute, the human side of delivering care, augmenting physician sensibilities of the human system over replacing the physician. Error rates of LLMs for complex diagnoses need to be constantly validated against medical benchmarks\cite{Singh2024}. Deviating from the norm could result in a misdiagnosis. Moreover, the models also need to be kept up-to-date through regular checking and flagging. 

That commitment to multidisciplinary work among technologists, clinicians, and regulators is also crucial to using LLMs in healthcare effectively. From conceiving AI tools that are both architecture- and design-wise solid and clinically useful to providing right-size validation and ensuring ethical use, stringent workflows across disciplines are necessary. Testing each model goes beyond mere optimization of technical performance to confirm that the tool is capable of effectively helping clinicians in actual practice\cite{10.1145/1579114.1579173}.

In conclusion, incorporating LLMs into healthcare is a promising avenue for improving diagnosis and patient care. However, to address these challenges, we must implement careful planning and robust processes. Doing so will help to make the most of AI in healthcare and improve outcomes for patients and the medical system.

\subsection{Limitations of the Study}

The LLMs in this study were primarily evaluated on common, less complex illnesses, which do not fully represent the broader, more challenging aspects of diagnosing chronic or severe conditions. Chronic diseases often involve complex symptoms that are difficult to interpret without a comprehensive understanding of an individual's medical history and additional diagnostic tests. Moreover, the reliance on text analysis in our study ignores the multimodal nature of traditional medical diagnostics, which often include visual elements like scans, detailed patient histories, and physical examinations. Future improvements in AI models for healthcare should aim to incorporate these various data types to provide more accurate and holistic diagnoses.

\subsection{Future Research Directions}

Future research should focus on creating LLMs that can analyze various types of medical information beyond just text. This includes integrating visual data from scans and other medical images to create more comprehensive diagnostic tools. Also, developing AI systems that can understand and process information across different languages and cultural contexts will be crucial for their global applicability. It’s also important to ensure these systems are developed with strict adherence to ethical standards, particularly regarding patient privacy. Exploring ways to securely integrate AI into healthcare while respecting patient confidentiality will be essential. Additionally, putting these AI models into real-world clinical settings to see how they perform and what impact they have on healthcare efficiency and patient outcomes will provide critical insights into their practical value and limitations.

 This research demonstrates the significant potential and challenges of using advanced AI models in healthcare diagnostics. With careful development and ethical considerations, these tools could greatly enhance the ability to diagnose and treat patients more efficiently and accurately. Continued exploration and improvement of these technologies could lead to their successful integration into everyday medical practice, benefiting both healthcare providers and patients.

\section{Conclusion}

The exploration into the capabilities of Large Language Models (LLMs) like GPT-4, Gemini, and GPT-3.5 has culminated in a comprehensive understanding of their potential to enhance digital diagnostics in healthcare. This study has rigorously evaluated the performance of these models, shedding light on their strengths in identifying symptoms of common illnesses and their limitations.
The core findings demonstrate that LLMs like GPT-4 offer considerable promise in processing medical language with high accuracy. They herald a significant step forward in providing immediate, accessible healthcare guidance. The specialized Gemini model’s remarkable precision points towards the feasibility of creating niche, domain-focused LLMs that can provide precise diagnostic support. GPT-3.5, while slightly overshadowed by its successors, still displays commendable capabilities, indicative of the rapid advancements within the field of AI in healthcare.These results reinforce the transformative potential of LLMs in digital diagnostics, suggesting that they can complement conventional diagnostic methods, enhancing the quality and accessibility of patient care. However, the journey from potential to actualized utility in a clinical setting will require overcoming substantial hurdles, including the integration of multimodal data, ethical considerations, and ensuring adherence to stringent healthcare standards.The limitations of this study, predominantly its scope restricted to textual analysis and the manual evaluation process, set a clear directive for future research. The next phase should aim to expand the capabilities of LLMs beyond text, incorporating visual and empirical data to align closely with comprehensive clinical diagnostics. Moreover, the study emphasizes the necessity to build ethically aligned, culturally sensitive, and linguistically diverse LLMs to serve global healthcare needs effectively.
In the quest to harness AI for healthcare, LLMs emerge not as standalone solutions but as part of a collaborative toolset augmenting the expertise of medical professionals. As research progresses, it will be paramount to embed these models within real-world clinical workflows to fully assess their practicality and reliability. With continued development and responsible implementation, LLMs are poised to play a pivotal role in shaping the future of healthcare, making diagnostics more accessible, accurate, and patient-centric.

\section*{Acknowledgments}
Thank you Dr. Aditi Singh as mentor for helping me in the process of experiment, writing the paper. 

\section*{Data availability}
The dataset associated in this manuscript, sourced from reputable medical institutions including the CDC, WHO, Mayo Clinic, Cleveland Clinic, and Johns Hopkins Hospital, are publicly available, focusing on common conditions to evaluate the diagnostic capabilities of Large Language Models (LLMs).

\section*{Declaration conflict of interest}
The author declares no conflict of interest.


\nocite{*}

\printbibliography

@article{mesko2017digital,
  title={The role of artificial intelligence in precision medicine},
  author={Mesk{\'o}, Bertalan and Het{\'e}nyi, Georgina and Győrffy, Zsuzsanna},
  journal={Expert review of precision medicine and drug development},
  volume={2},
  number={5},
  pages={239--241},
  year={2017},
  publisher={Taylor \& Francis}
}

@article{yu2020artificial,
  title={Artificial intelligence in healthcare: A critical analysis of the legal and ethical implications},
  author={Yu, Katherine and Beam, Andrew L and Kohane, Isaac S},
  journal={International journal of medical informatics},
  volume={141},
  pages={104431},
  year={2020},
  publisher={Elsevier}
}

@inproceedings{yuan2021improving,
  title={Improving patient-trial matching with a data-driven approach},
  author={Yuan, Jiayi and others},
  booktitle={International Conference on Medical Informatics},
  year={2021}
}

@article{kusa2021prompting,
  title={“Dr LLM, what do I have?”: The impact of user beliefs and prompt formulation on health diagnoses},
  author={Kusa, Wojciech and Mosca, Edoardo and Lipani, Aldo},
  journal={Journal of Medical Internet Research},
  year={2021},
  publisher={JMIR Publications Inc., Toronto, Canada}
}

@misc{openai2023gpt4,
    title={GPT-4 Technical Report},
    author={OpenAI},
    year={2023},
    eprint={2303.08774},
    archivePrefix={arXiv},
    primaryClass={cs.CL}
}

@misc{geminiteam2024gemini,
      title={Gemini: A Family of Highly Capable Multimodal Models}, 
      author={Gemini Team},
      year={2024},
      eprint={2312.11805},
      archivePrefix={arXiv},
      primaryClass={cs.CL}
}

@misc{dhakal2024gpt4s,
      title={GPT-4's assessment of its performance in a USMLE-based case study}, 
      author={Uttam Dhakal and Aniket Kumar Singh and Suman Devkota and Yogesh Sapkota and Bishal Lamichhane and Suprinsa Paudyal and Chandra Dhakal},
      year={2024},
      eprint={2402.09654},
      archivePrefix={arXiv},
      primaryClass={cs.AI}
}

@article{Singh2024,
  author  = {Aniket Kumar Singh and Bishal Lamichhane and Suman Devkota and Uttam Dhakal and Chandra Dhakal},
  title   = {Do Large Language Models Show Human-like Biases? Exploring Confidence—Competence Gap in {AI}},
  journal = {Information},
  year    = {2024},
  volume  = {15},
  number  = {2},
  pages   = {92},
  doi     = {10.3390/info15020092},
  url     = {https://doi.org/10.3390/info15020092}
}

@article{baricparker2020patient,
  author       = {J Baric-Parker and E.E. Anderson},
  title        = {Patient Data-Sharing for AI: Ethical Challenges, Catholic Solutions},
  journal      = {The Linacre Quarterly},
  year         = {2020},
  volume       = {87},
  number       = {4},
  pages        = {471-481},
  doi          = {10.1177/0024363920922690}
}

@phdthesis{humphrey2021privacy,
  author       = {Barry A. Humphrey},
  title        = {Data Privacy vs. Innovation: A Quantitative Analysis of Artificial Intelligence in Healthcare and Its Impact on HIPAA regarding the Privacy and Security of Protected Health Information},
  school       = {Robert Morris University},
  year         = {2021},
  type         = {PhD dissertation},
  publisher    = {ProQuest Dissertations Publishing},
  number       = {28549541}
}

@Article{healthcare11060887,
AUTHOR = {Sallam, Malik},
TITLE = {ChatGPT Utility in Healthcare Education, Research, and Practice: Systematic Review on the Promising Perspectives and Valid Concerns},
JOURNAL = {Healthcare},
VOLUME = {11},
YEAR = {2023},
NUMBER = {6},
ARTICLE-NUMBER = {887},
URL = {https://www.mdpi.com/2227-9032/11/6/887},
PubMedID = {36981544},
ISSN = {2227-9032},
DOI = {10.3390/healthcare11060887}
}

@inproceedings{10.1145/3544548.3581503,
author = {Jo, Eunkyung and Epstein, Daniel A. and Jung, Hyunhoon and Kim, Young-Ho},
title = {Understanding the Benefits and Challenges of Deploying Conversational AI Leveraging Large Language Models for Public Health Intervention},
year = {2023},
isbn = {9781450394215},
publisher = {Association for Computing Machinery},
address = {New York, NY, USA},
url = {https://doi.org/10.1145/3544548.3581503},
doi = {10.1145/3544548.3581503},
booktitle = {Proceedings of the 2023 CHI Conference on Human Factors in Computing Systems},
articleno = {18},
numpages = {16},
location = {<conf-loc>, <city>Hamburg</city>, <country>Germany</country>, </conf-loc>},
series = {CHI '23}
}

@misc{lai2023psyllm,
      title={Psy-LLM: Scaling up Global Mental Health Psychological Services with AI-based Large Language Models}, 
      author={Tin Lai and Yukun Shi and Zicong Du and Jiajie Wu and Ken Fu and Yichao Dou and Ziqi Wang},
      year={2023},
      eprint={2307.11991},
      archivePrefix={arXiv},
      primaryClass={cs.CL}
}

@misc{ghosh2023clipsyntel,
      title={CLIPSyntel: CLIP and LLM Synergy for Multimodal Question Summarization in Healthcare}, 
      author={Akash Ghosh and Arkadeep Acharya and Raghav Jain and Sriparna Saha and Aman Chadha and Setu Sinha},
      year={2023},
      eprint={2312.11541},
      archivePrefix={arXiv},
      primaryClass={cs.AI}
}

@article{webster2023sixways,
  author    = {Webster, P.},
  title     = {Six Ways Large Language Models are Changing Healthcare},
  journal   = {Nat Med},
  year      = {2023},
  volume    = {29},
  pages     = {2969--2971},
  doi       = {10.1038/s41591-023-02700-1},
  url       = {https://doi.org/10.1038/s41591-023-02700-1}
}

@misc{abbasian2024conversational,
      title={Conversational Health Agents: A Personalized LLM-Powered Agent Framework}, 
      author={Mahyar Abbasian and Iman Azimi and Amir M. Rahmani and Ramesh Jain},
      year={2024},
      eprint={2310.02374},
      archivePrefix={arXiv},
      primaryClass={cs.CL}
}

@inproceedings{10.1145/3582515.3609536,
author = {Montagna, Sara and Ferretti, Stefano and Klopfenstein, Lorenz Cuno and Florio, Antonio and Pengo, Martino Francesco},
title = {Data Decentralisation of LLM-Based Chatbot Systems in Chronic Disease Self-Management},
year = {2023},
isbn = {9798400701160},
publisher = {Association for Computing Machinery},
address = {New York, NY, USA},
url = {https://doi.org/10.1145/3582515.3609536},
doi = {10.1145/3582515.3609536},
booktitle = {Proceedings of the 2023 ACM Conference on Information Technology for Social Good},
pages = {205–212},
numpages = {8},
location = {Lisbon, Portugal},
series = {GoodIT '23}
}

@article{peng2024evaluating,
  author    = {Peng, W. and Feng, Y. and Yao, C. and others},
  title     = {Evaluating AI in Medicine: A Comparative Analysis of Expert and ChatGPT Responses to Colorectal Cancer Questions},
  journal   = {Sci Rep},  % Abbreviation for Scientific Reports
  year      = {2024},
  volume    = {14},
  pages     = {2840},
  doi       = {10.1038/s41598-024-52853-3},
  url       = {https://doi.org/10.1038/s41598-024-52853-3}
}

@article{10.1145/3643540,
author = {Xu, Xuhai and Yao, Bingsheng and Dong, Yuanzhe and Gabriel, Saadia and Yu, Hong and Hendler, James and Ghassemi, Marzyeh and Dey, Anind K. and Wang, Dakuo},
title = {Mental-LLM: Leveraging Large Language Models for Mental Health Prediction via Online Text Data},
year = {2024},
issue_date = {March 2024},
publisher = {Association for Computing Machinery},
address = {New York, NY, USA},
volume = {8},
number = {1},
url = {https://doi.org/10.1145/3643540},
doi = {10.1145/3643540},
journal = {Proc. ACM Interact. Mob. Wearable Ubiquitous Technol.},
month = {mar},
articleno = {31},
numpages = {32}
}

@misc{kim2024healthllm,
      title={Health-LLM: Large Language Models for Health Prediction via Wearable Sensor Data}, 
      author={Yubin Kim and Xuhai Xu and Daniel McDuff and Cynthia Breazeal and Hae Won Park},
      year={2024},
      eprint={2401.06866},
      archivePrefix={arXiv},
      primaryClass={cs.CL}
}

@misc{jin2024healthllm,
      title={Health-LLM: Personalized Retrieval-Augmented Disease Prediction System}, 
      author={Mingyu Jin and Qinkai Yu and Dong Shu and Chong Zhang and Lizhou Fan and Wenyue Hua and Suiyuan Zhu and Yanda Meng and Zhenting Wang and Mengnan Du and Yongfeng Zhang},
      year={2024},
      eprint={2402.00746},
      archivePrefix={arXiv},
      primaryClass={cs.CL}
}

@article{ullah2024challenges,
  author    = {Ullah, E. and Parwani, A. and Baig, M. M. and others},
  title     = {Challenges and Barriers of Using Large Language Models (LLM) Such as ChatGPT for Diagnostic Medicine with a Focus on Digital Pathology – A Recent Scoping Review},
  journal   = {Diagn Pathol},
  year      = {2024},
  volume    = {19},
  pages     = {43},
  doi       = {10.1186/s13000-024-01464-7},
  url       = {https://doi.org/10.1186/s13000-024-01464-7}
}

@article{Choudhury_2024,
   title={Large Language Models and User Trust:  Focus on Healthcare  (Preprint)},
   ISSN={1438-8871},
   url={http://dx.doi.org/10.2196/56764},
   DOI={10.2196/56764},
   journal={Journal of Medical Internet Research},
   publisher={JMIR Publications Inc.},
   author={Choudhury, Avishek and Chaudhry, Zaira},
   year={2024},
   month=jan }

@article{osti_10448809,
place = {Country unknown/Code not available}, title = {LLM for Patient-Trial Matching: Privacy-Aware Data Augmentation Towards Better Performance and Generalizability}, url = {https://par.nsf.gov/biblio/10448809}, journal = {American Medical Informatics Association (AMIA) Annual Symposium}, author = {Yuan J and Tang R and Jiang X and Hu X}, }

@inproceedings{10.1145/1579114.1579173,
author = {Frantzidis, Christos A. and Bamidis, Panagiotis D.},
title = {Description and future trends of ICT solutions offered towards independent living: the case of LLM project},
year = {2009},
isbn = {9781605584096},
publisher = {Association for Computing Machinery},
address = {New York, NY, USA},
url = {https://doi.org/10.1145/1579114.1579173},
doi = {10.1145/1579114.1579173},
booktitle = {Proceedings of the 2nd International Conference on PErvasive Technologies Related to Assistive Environments},
articleno = {59},
numpages = {8},
location = {Corfu, Greece},
series = {PETRA '09}
}

@article{batsis2017leanmass,
  author    = {Batsis, John A. and Mackenzie, Todd A. and Emeny, Rebecca T. and Lopez-Jimenez, Francisco and Bartels, Stephen J.},
  title     = {Low Lean Mass With and Without Obesity, and Mortality: Results From the 1999–2004 National Health and Nutrition Examination Survey},
  journal   = {The Journals of Gerontology: Series A},
  year      = {2017},
  volume    = {72},
  issue     = {10},
  pages     = {1445--1451},
  doi       = {10.1093/gerona/glx002},
  url       = {https://doi.org/10.1093/gerona/glx002}
}

@misc{cui2024multimodal,
      title={Multimodal Fusion of EHR in Structures and Semantics: Integrating Clinical Records and Notes with Hypergraph and LLM}, 
      author={Hejie Cui and Xinyu Fang and Ran Xu and Xuan Kan and Joyce C. Ho and Carl Yang},
      year={2024},
      eprint={2403.08818},
      archivePrefix={arXiv},
      primaryClass={cs.LG}
}

@misc{chiu2024computational,
      title={A Computational Framework for Behavioral Assessment of LLM Therapists}, 
      author={Yu Ying Chiu and Ashish Sharma and Inna Wanyin Lin and Tim Althoff},
      year={2024},
      eprint={2401.00820},
      archivePrefix={arXiv},
      primaryClass={cs.CL}
}

@Article{electronics12132814,
AUTHOR = {de Curtò, J. and de Zarzà, I. and Roig, Gemma and Cano, Juan Carlos and Manzoni, Pietro and Calafate, Carlos T.},
TITLE = {LLM-Informed Multi-Armed Bandit Strategies for Non-Stationary Environments},
JOURNAL = {Electronics},
VOLUME = {12},
YEAR = {2023},
NUMBER = {13},
ARTICLE-NUMBER = {2814},
URL = {https://www.mdpi.com/2079-9292/12/13/2814},
ISSN = {2079-9292},
DOI = {10.3390/electronics12132814}
}

@article{baharudin2022lipidlowering,
  author    = {Baharudin, N. and Mohamed-Yassin, M. S. and Daher, A. M. and others},
  title     = {Prevalence and factors associated with lipid-lowering medications use for primary and secondary prevention of cardiovascular diseases among Malaysians: the REDISCOVER study},
  journal   = {BMC Public Health},
  year      = {2022},
  volume    = {22},
  pages     = {228},
  doi       = {10.1186/s12889-022-12595-1},
  url       = {https://doi.org/10.1186/s12889-022-12595-1}
}

@INPROCEEDINGS{10427865,
  author={Singh, Aditi and Ehtesham, Abul and Mahmud, Saifuddin and Kim, Jong-Hoon},
  booktitle={2024 IEEE 14th Annual Computing and Communication Workshop and Conference (CCWC)}, 
  title={Revolutionizing Mental Health Care through LangChain: A Journey with a Large Language Model}, 
  year={2024},
  volume={},
  number={},
  pages={0073-0078},
  doi={10.1109/CCWC60891.2024.10427865}}

\begin{figure}[h]
    \centering
    \includegraphics[width=\textwidth]{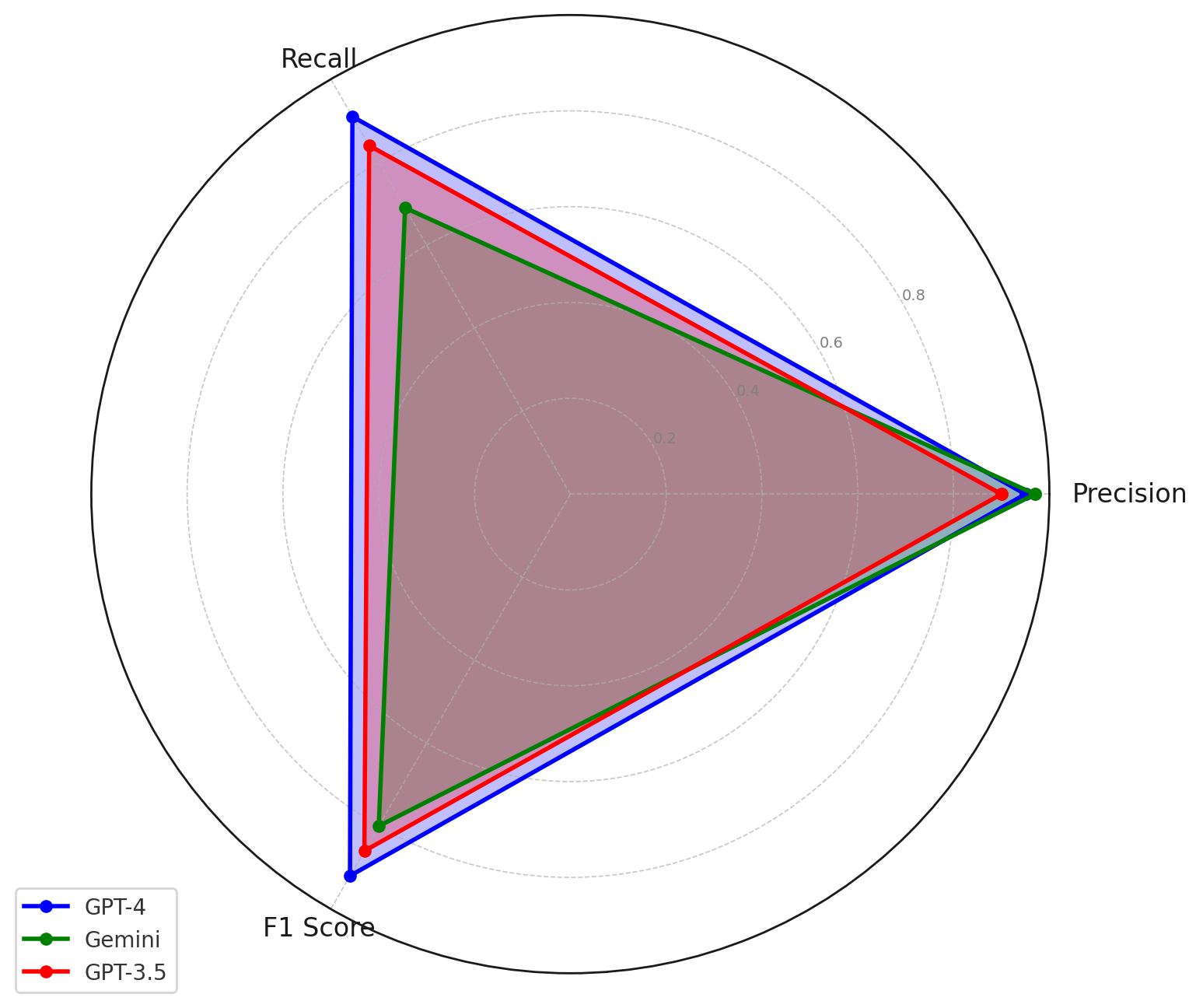}
    \caption{Performance Metrics}
    \label{fig:Performance Metrics}
\end{figure}

\end{document}